\PassOptionsToPackage{hyperfootnotes=false}{hyperref}
\documentclass[times,review,10pt]{elsarticle}

\usepackage[
  top=4.3cm,
  right=4.8cm,
  bottom=4.3cm,
  left=4.8cm
]{geometry}

\usepackage{amsmath}
\usepackage{amssymb}
\usepackage{algorithm}
\usepackage{algorithmic}
\usepackage[caption=false,font=small]{subfig}
\usepackage{textcomp}
\usepackage{url}
\usepackage{dsfont}
\usepackage{verbatim}
\usepackage{threeparttable}
\usepackage{diagbox}
\usepackage{booktabs}
\usepackage{makecell}
\usepackage{multirow}
\usepackage[dvipsnames]{xcolor}
\usepackage{hyperref}
\usepackage{tabularx}
\usepackage{array}

\hypersetup{
  breaklinks=true,
  colorlinks=true,
  urlcolor=blue,
  citecolor=blue,
  linkcolor=BrickRed
}

\journal{Pattern Recognition}
\biboptions{sort&compress}

\newcommand{\rui}[1]{{\color{black}#1}}

\newcommand{\nodata}{\mbox{--}}
\newcommand{\methodname}{MMGNet}

\begin{document}

\begin{frontmatter}

\title{MMGNet: Multi-Modal 6-DoF Grasp Synthesis for Cluttered Scenes}

\author[1]{Rui Cao}
\ead{rcao@mae.cuhk.edu.hk}
\author[2]{Jia-Hui Pan}
\author[1]{Yichuan Li}
\author[1]{Yunfan Yu}
\author[2]{Chi-Wing Fu}
\author[1,3]{Jiangliu Wang}
\author[1]{Yun-Hui Liu\corref{cor1}}
\ead{yhliu@mae.cuhk.edu.hk}

\affiliation[1]{organization={Department of Mechanical and Automation Engineering, The Chinese University of Hong Kong},
  city={Hong Kong},
  country={China}}
\affiliation[2]{organization={Department of Computer Science and Engineering, The Chinese University of Hong Kong},
  city={Hong Kong},
  country={China}}
\affiliation[3]{organization={Sino-German College of Intelligent Manufacturing, Shenzhen Technology University},
  city={Shenzhen},
  country={China}}
\cortext[cor1]{Corresponding author}

\begin{abstract}
Grasping is a foundational capability in robotic manipulation.
Scene-level 6-DoF grasp synthesis relies on point clouds for metric geometry, yet real-world point observations are often noisy, incomplete, and over-smoothed.
RGB images provide object-region, boundary, and appearance cues, yet two bottlenecks limit their effective use in scene representation and local grasp reasoning.
We present \methodname, a Multi-Modal Grasp Synthesis Network that addresses these bottlenecks through hierarchical image-to-3D feature injection and rotational graspness modeling.
Hierarchical injection integrates multi-scale image features throughout 3D representation formation, while the rotational graspness jointly predicts the grasp approach direction and in-plane rotation before organizing local image-informed features through grasp-shaded grouping.
Ablation studies show that these two modules provide complementary gains.
\methodname\ achieves a mean AP of 60.34 on GraspNet-1Billion with RealSense, surpassing all compared methods.
In real-world experiments with adversarial grasp objects, \methodname\ improves the grasp success rate over AnyGrasp by 8.46 percentage points.
\emph{Code, models, and data will be released upon publication.}
\end{abstract}


\end{frontmatter}

\section{Introduction}
\label{sec:introduction}
Robotic grasping supports downstream tasks such as bin picking and packing~\cite{li2022sim,pan2024ppn}, object rearrangement~\cite{danielczuk2021object}, and tool manipulation~\cite{laili2022custom}.
Scene-level 6-DoF grasp synthesis predicts executable gripper poses for objects with diverse shapes, orientations, and occlusions.
Most existing scene-level methods rely primarily on point clouds to represent the metric geometry required for 6-DoF grasp prediction~\cite{graspnet,anygrasp,ma2023towards,graspness}.
However, real-world point clouds often suffer from measurement noise, local incompleteness, and over-smoothing, while RGB images provide complementary cues about object regions, boundaries, and appearance.
Accordingly, several grasp synthesis methods have incorporated image information to improve grasp prediction~\cite{gou2021rgb,liu2023joint,xie2026rethinking}.

Nevertheless, two bottlenecks limit the effective use of image information in scene representation and local grasp reasoning.
\rui{
First, some existing methods incorporate image cues through late feature fusion~\cite{liu2023joint} or during local-region processing~\cite{xie2026rethinking}.
Consequently, multi-scale image context plays a limited role during 3D scene representation formation.
We therefore argue that such image information should be integrated throughout this process.}
Second, even when image cues are embedded in the scene representation, the local representation used for detailed grasp prediction remains too coarse to capture grasp-quality differences across gripper rotations.
Existing pipelines~\cite{graspness,anygrasp,ma2023towards} first use grasp-quality scores to select promising grasp centers and approach directions, aggregate nearby 3D features, and only then predict the gripper rotation around the selected direction.
Consequently, grasps that share the same center and approach direction but differ in gripper rotation are evaluated from the same local representation.
Thus, the embedded image cues are not organized according to the candidate rotation, limiting their ability to capture the corresponding grasp-quality differences.

We present the Multi-Modal Grasp Synthesis Network (\methodname), a single-view scene-level framework designed to address these two bottlenecks.
To address the first, \methodname\ introduces hierarchical image-to-3D feature injection, which aligns multi-scale image features with scene points and injects them into successive stages of a sparse 3D backbone, allowing image context to shape the 3D representation throughout the backbone.
To address the second, \methodname\ introduces rotational graspness modeling to jointly predict the grasp approach direction and in-plane rotation before local aggregation.
The selected rotation then guides grasp-shaded grouping, which gathers local image-informed 3D features in the corresponding gripper frame.
Together, these designs make grasp-quality differences across rotations explicit before local analysis and organize the resulting local features according to the selected gripper orientation.

Given an RGB-D observation, \methodname\ first extracts a multi-scale image feature pyramid and establishes pixel--point correspondence with the full-scene point cloud.
The aligned image features are injected into multiple sparse-encoder stages to form an RGB-informed 3D scene representation while retaining efficient metric neighborhood aggregation.
A foreground score then guides the sampling of spatially distributed grasp seeds.
For each seed, rotational graspness selects a joint view--in-plane rotation, after which grasp-shaded grouping aggregates local features in the corresponding gripper frame.
Finally, the operation head predicts approach depth and width of the gripper to produce the 6-DoF grasps.

In GraspNet-1Billion~\cite{graspnet} dataset, \methodname\ achieves a state-of-the-art mean Average Precision (AP) of 60.34 in RealSense scenes and a competitive mean AP of 50.72 in Kinect scenes.
Ablation studies show that hierarchical image-to-3D feature injection improves the point-only Rotational model by 5.39 AP points, while the rotational graspness model outperforms the View model by 8.39 points under the same hierarchical injection.
The two modules exhibit a positive interaction of 2.19 AP points, confirming their complementarity.
Cross-modal representation analysis further reveals stronger image--3D correspondence across backbone stages than early and late fusion.
Point- and RGB-corruption experiments show that image-informed 3D features complement noisy or incomplete point observations and characterize how different fusion strategies respond to unreliable RGB.
In real-robot experiments, \methodname\ improves the grasp success rate over AnyGrasp~\cite{anygrasp} by 8.46 percentage points on adversarial grasp objects~\cite{wang2019adversarial}. Our main contributions are:
\begin{itemize}
    \item We develop \methodname, a scene-level multi-modal framework for single-view 6-DoF grasp synthesis.

    \item We introduce hierarchical image-to-3D feature injection, which integrates multi-scale image cues throughout 3D representation formation.

    \item We introduce a rotational graspness model that scores joint view--in-plane rotations and uses the selected rotation to organize image-informed local features through grasp-shaded grouping.

    \item \methodname\ achieves state-of-the-art performance on GraspNet-1Billion and demonstrates its applicability in real-world robotic grasping.

    \item Extensive ablations and analyses validate the proposed modules across fusion strategies, grasp models, representation properties, and point- and RGB-corrupted observations.
\end{itemize}

\section{Related Work}
\subsection{6-DoF Grasp Synthesis for Cluttered Scenes}
Grasp synthesis predicts executable gripper poses from visual observations.
Early image-based methods mainly considered 4-DoF top-down grasps, represented as planar rectangles in RGB or RGB-D images~\cite{depierre2018jacquard,ainetter2021end,jiang2011efficient}.
Although efficient, this formulation restricts the gripper approach direction.
Six-DoF grasp synthesis removes this restriction and enables more flexible grasping in cluttered scenes.

\noindent\textbf{Point-cloud-based methods.}
Most scene-level 6-DoF methods operate on depth-derived point clouds.
Zeng~\textit{et al.}~\cite{zeng2018robotic} predict dense affordances for robotic manipulation, while Fang~\textit{et al.}~\cite{graspnet} lift dense grasp prediction into 3D scene points.
GSNet~\cite{graspness} introduces graspness to identify promising regions before detailed grasp prediction, reducing the scene-level search space.
Subsequent methods improve candidate selection through adaptive instance sampling~\cite{ma2023towards}, graspable-region prediction~\cite{chen2023grasp}, or interest-point selection~\cite{he2024interest}.
AnyGrasp~\cite{anygrasp} further incorporates object center-of-mass information to improve grasp stability.
\rui{These methods progressively improve candidate quality and computational efficiency, but remain primarily dependent on partial and potentially noisy point observations.}

\noindent\textbf{RGB-D and richer observations.}
Several methods incorporate image information or additional geometric context into 6-DoF grasp synthesis.
RGB Matters~\cite{gou2021rgb} processes RGB and depth separately for gripper orientation prediction and analytic grasp search.
Liu~\textit{et al.}~\cite{liu2023joint} \rui{combine independently encoded image and point features through late fusion}, while Qiu~\textit{et al.}~\cite{qiu2023multi} concatenate range-image and point-cloud features to recover boundary information missing from depth observations.
CenterGrasp~\cite{chisari2024centergrasp} combines object-aware shape reconstruction with grasp estimation, and ZeroGrasp~\cite{zerograsp} jointly reconstructs complete 3D geometry and predicts grasps from a single RGB-D observation using occlusion and inter-object reasoning.
FlexLoG~\cite{xie2026rethinking} combines flexible guidance with a local grasp model to support both scene-level and target-oriented grasping.
Zhu~\textit{et al.}~\cite{zhu2026crossview} improve robustness to occlusion by post-fusing an auxiliary RGB-D view and aggregating cross-view geometry around grasp candidates.

These works exploit richer visual information through late feature fusion, shape reconstruction, local-region modeling, or cross-view aggregation.
\methodname\ follows the scene-level point-cloud paradigm and integrates multi-scale image information into the 3D scene representation.
\rui{The resulting image-informed representation is then coupled with rotational graspness and grasp-shaded grouping for local grasp prediction across different gripper rotations.}

\subsection{Image-to-3D Fusion for 3D Perception}
Image-to-3D fusion has been explored at input, intermediate, and output levels in 3D object detection~\cite{wang2021pointaugmenting,li2022deepfusion}, segmentation~\cite{el2019rgb}, and scene understanding~\cite{robert2022learning,peng2023openscene}.
Multi-level 2D--3D interaction has also been studied for scene understanding and 6D object pose estimation~\cite{hu2021bidirectional,he2021ffb6d}.
These studies demonstrate the benefit of integrating image information at multiple levels of 3D feature learning.
Within this line of work, \methodname\ introduces hierarchical image-to-3D injection into a sparse scene-level grasp backbone.
Multi-scale image features are aligned with scene points and injected into successive backbone stages, allowing image information to participate throughout 3D scene representation formation.

\begin{figure*}[t]
    \centering
    \includegraphics[width=\linewidth]{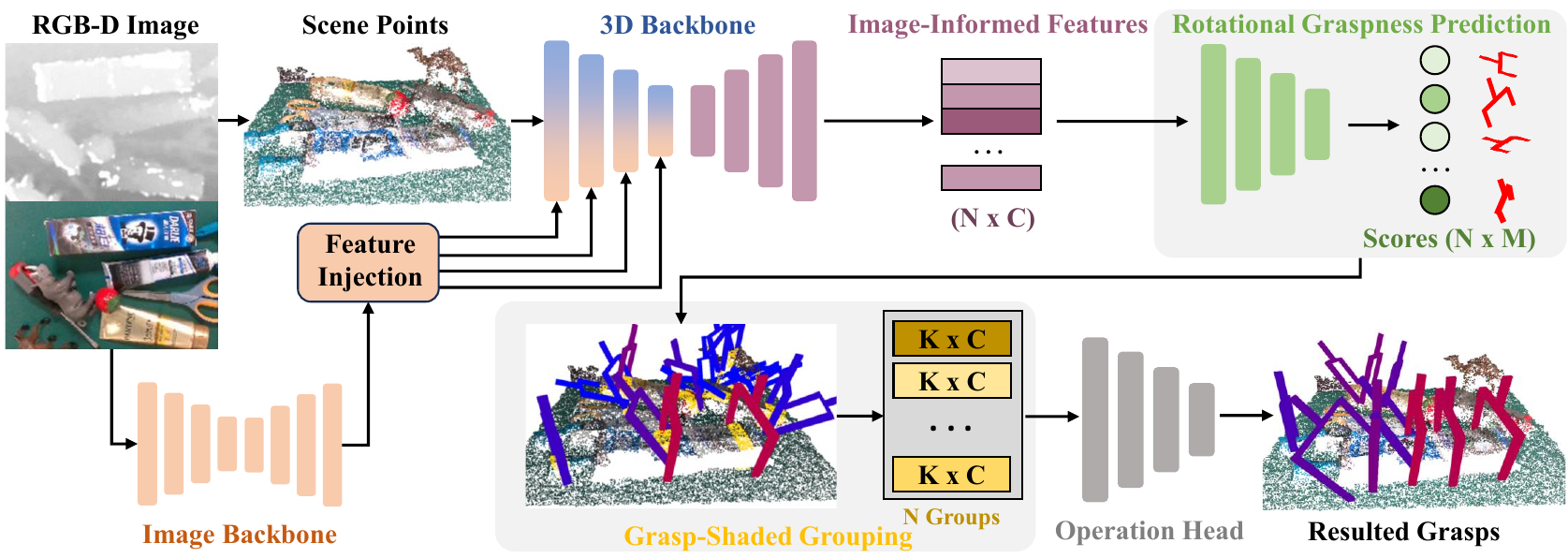}
    \caption{Illustration of \methodname. Given a single-view RGB-D observation, the image backbone extracts dense image features, which are aligned with the scene points through pixel--point correspondence and injected into successive stages of the 3D backbone to form image-informed 3D features. Rotational graspness prediction then scores joint view--in-plane rotations at candidate grasp centers. Based on the selected rotations, grasp-shaded grouping aggregates local image-informed features within the corresponding gripper frames. Finally, the operation head predicts the remaining depth-wise grasp scores and gripper widths to produce the final 6-DoF grasps.}
     \label{fig:framework}
\end{figure*}

\section{Method}
\subsection{Problem Formulation and Pipeline Overview}
Grasp synthesis predicts a set of executable grasps for a robot arm to pick up objects in a cluttered scene.
Given a single-view RGB-D observation, a grasp synthesis method predicts a set of 6-DoF grasp poses $\mathcal{G}=\{\mathbf{g}_r\}_{r=1}^{N_g}$.
A grasp pose is defined as $\mathbf{g}=(\mathbf{R}, \mathbf{t}, \omega)$, where $\mathbf{R} \in \mathbb{R}^{3 \times 3}$, $\mathbf{t} \in \mathbb{R}^{3}$, and $\omega \in \mathbb{R}_{+}$ denote the gripper rotation, translation, and width, respectively.

We present \methodname, a scene-level 6-DoF grasp synthesis network built around hierarchical image-to-3D feature injection and a rotational graspness model.
\methodname\ extracts dense multi-scale image features, establishes pixel--point correspondence, and injects the aligned image features at multiple encoder stages of a sparse 3D backbone.
This allows visual cues to participate throughout 3D representation formation rather than serving only as input initialization or late augmentation.
From the resulting image-informed 3D scene representation, a point-wise objectness head identifies foreground object regions, and Farthest Point Sampling (FPS) selects spatially distributed seed points for grasp prediction.
\methodname\ then applies the rotational graspness model, which resolves the joint view--in-plane rotation for each seed and aggregates local features within the corresponding grasp-shaded region to predict depth-wise grasp scores and gripper widths.

\begin{figure}[t]
    \centering
    \includegraphics[width=\linewidth]{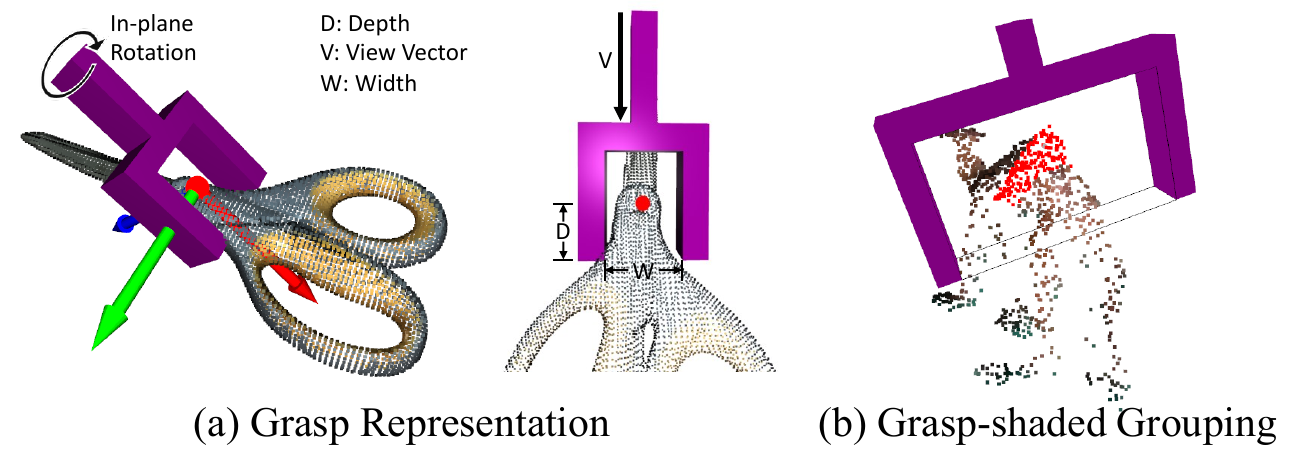}
    \caption{(a) Grasp representation in \methodname, showing the grasp coordinate frame with the red point marking the grasp origin. (b) Grasp-shaded grouping, where the black cuboid denotes the grouping region and the red points denote the grouped scene points.}
    \label{fig:grasp_representation}
\end{figure}

\subsection{Hierarchical Image-to-3D Feature Injection}
\label{sec:method}
To form an image-informed 3D scene representation, \methodname\ aligns multi-scale image features with scene points and injects them at successive stages of a sparse 3D backbone.
This hierarchical injection allows RGB information to participate throughout 3D representation formation.
An objectness head then identifies foreground regions, from which FPS selects spatially distributed seeds for grasp prediction.

\noindent\textbf{Multi-Scale Image Feature Extraction.}
Given the full RGB image $\mathbf{I}$, we use a convolutional neural network (CNN) $\Psi$ to extract a multi-scale dense image feature pyramid:
\begin{equation}
\{\mathbf{F}^{(1)}_{\mathrm{img}}, \mathbf{F}^{(2)}_{\mathrm{img}}, \mathbf{F}^{(4)}_{\mathrm{img}}, \mathbf{F}^{(8)}_{\mathrm{img}}, \mathbf{F}^{(16)}_{\mathrm{img}}\} = \Psi(\mathbf{I}),
\end{equation}
where the superscript denotes the feature stride relative to the input image.
We use ResNet~\cite{he2016deep} and PSPNet~\cite{zhao2017pyramid} to encode dense visual evidence at multiple spatial scales.
The resulting feature pyramid provides visual features for the hierarchical image-to-3D injection described next.

\noindent\textbf{Hierarchical Image-to-3D Feature Injection.}
To transfer image evidence into the 3D scene representation, we first establish the correspondence between each scene point and its image pixel.
Given a scene point $\mathbf{p}_i=(x_i,y_i,z_i)^\top \in \mathcal{P}$ in the camera coordinate system, its corresponding pixel $(u_i,v_i)$ is obtained using the pinhole camera model with camera intrinsic matrix $\mathbf{K} \in \mathbb{R}^{3 \times 3}$:
\begin{equation}
[u_i, v_i, 1]^\top = \frac{1}{z_i} \mathbf{K} \mathbf{p}_i .
\end{equation}

For each feature stride $s \in \{1,2,4,8,16\}$, we gather the per-point image feature from the corresponding feature map:
\begin{equation}
\mathbf{f}^{(s)}_i =
\mathbf{F}^{(s)}_{\mathrm{img}}
\left(
\left\lfloor \frac{u_i}{s} \right\rfloor,
\left\lfloor \frac{v_i}{s} \right\rfloor
\right).
\end{equation}

Rather than injecting visual cues only before sparse 3D encoding or after the 3D hierarchy has formed, \methodname\ injects them throughout the hierarchical sparse encoder.
At encoder stage $s$, let $\mathbf{x}^{(s)}_\ell$ denote the feature of the $\ell$-th active voxel, and let $\mathcal{Q}^{(s)}_\ell$ be the set of scene points assigned to that voxel.
We aggregate the aligned point-wise image features onto active sparse voxels by mean scattering:
\begin{equation}
\tilde{\mathbf{f}}^{(s)}_\ell =
\frac{1}{|\mathcal{Q}^{(s)}_\ell|}
\sum_{i \in \mathcal{Q}^{(s)}_\ell} \mathbf{f}^{(s)}_i .
\end{equation}

We concatenate the aggregated image feature with the sparse 3D feature and project the result back to the original channel dimension using a $1 \times 1$ sparse convolution:
\begin{equation}
\bar{\mathbf{x}}^{(s)}_\ell =
\Omega^{(s)}([\mathbf{x}^{(s)}_\ell,\tilde{\mathbf{f}}^{(s)}_\ell]),
\end{equation}
where $\Omega^{(s)}(\cdot)$ denotes the stage-specific sparse projection layer.
We apply this injection at encoder stages $s$, allowing dense visual cues to shape the 3D representation at successive spatial scales while preserving the active sparse voxel structure.
This hierarchical injection defines the image-to-3D interface of \methodname: it enriches the learned 3D features for grasp reasoning without modifying the point coordinates.

After sparse encoder--decoder processing, the image-informed voxel features are mapped back to the original scene points, yielding image-informed scene features $\mathbf{F}_{\mathrm{scene}} \in \mathbb{R}^{N_f \times C}$ for objectness prediction and grasp synthesis.
Given its strong performance across 3D tasks~\cite{rocco2020efficient,Schult23ICRA}, we employ MinkowskiEngine~\cite{choy20194d} to implement the sparse 3D backbone.

\noindent\textbf{Objectness-guided Seed Selection.}
From the image-informed scene features $\mathbf{F}_{\mathrm{scene}}$, a three-layer MLP predicts a point-wise objectness score $o_i$ for each full-scene point $\mathbf{p}_i$.
Here, objectness denotes a class- and instance-agnostic foreground probability indicating whether a scene point lies on a visible object surface within the workspace.
We then apply FPS to the predicted object points to select a compact set of $N$ seed points for grasp prediction.
This step suppresses background clutter and maintains spatial coverage over the predicted object regions, allowing the following grasp prediction modules to focus on grasp-relevant scene points.

\subsection{Rotational Graspness Model}
Following seed selection, the rotational graspness model predicts grasps through two sequential stages.
Rotational graspness prediction first selects a joint view--in-plane rotation for each seed, after which grasp-shaded grouping aggregates image-informed local feature in the corresponding gripper frame.
The operation head then predicts the remaining depth-wise grasp parameters.

\noindent\textbf{Rotational Graspness Prediction.}
Previous graspness-based methods predict view graspness in the first stage and leave in-plane rotation and depth to subsequent grasp prediction.
Let $\hat{q}_{n,v,a,d}$ denote the analytic grasp quality for seed $n$, view $v$, in-plane rotation $a$, and depth $d$.
With $A$ in-plane rotations and $D$ depth anchors, view graspness summarizes grasp quality over both dimensions:
\begin{equation}
\hat{s}^{\mathrm{view}}_{n,v}
=
\frac{1}{AD}
\sum_{a=1}^{A}
\sum_{d=1}^{D}
\mathds{1}(\hat{q}_{n,v,a,d}>c),
\end{equation}
where $c$ is the grasp-quality threshold.
Thus, different in-plane rotations under the same view share one graspness score, although subsequent grasp prediction must distinguish them.
Moreover, the in-plane rotation changes the gripper placement and therefore the local contact and collision geometry.

To retain this distinction before local aggregation, we define rotational graspness for each joint view--in-plane rotation:
\begin{equation}
\hat{s}^{\mathrm{rot}}_{n,v,a}
=
\frac{1}{D}
\sum_{d=1}^{D}
\mathds{1}(\hat{q}_{n,v,a,d}>c).
\end{equation}
Unlike view graspness, rotational graspness averages only over depth and assigns a separate graspness target to each $(v,a)$ pair.
\rui{
Assigning a separate target to each joint view--in-plane rotation encourages the pre-grouping image-informed seed features to retain grasp-quality differences among candidate rotations.
}
We normalize the ground-truth rotational graspness targets across seed points in each scene to the range $[0,1]$.

For implementation, we flatten the $V\times A$ view--in-plane rotation pairs into $M=VA$ rotation anchors and denote the predicted scores as
$\mathbf{S}_{\mathrm{rot}}=\{s_{n,m}\}\in\mathbb{R}^{N\times M}$.
An MLP predicts $\mathbf{S}_{\mathrm{rot}}$ from the image-informed seed features.
For each seed, we select the rotation anchor with the highest predicted graspness score:
\begin{equation}
(v_n^{*},a_n^{*})
=
\arg\max_{v,a}
s^{\mathrm{rot}}_{n,v,a}.
\end{equation}
The selected pair determines the gripper rotation
$\mathbf{R}_n=\mathbf{R}(v_n^{*},a_n^{*})$,
which is used by grasp-shaded grouping for subsequent local analysis.
The operation head then predicts depth-wise grasp scores and gripper widths for this selected rotation.

\noindent\textbf{Grasp-shaded Grouping.}
As the second stage of the rotational graspness model, grasp-shaded grouping constructs a gripper-aligned local support region after the joint view--in-plane rotation is selected.
For the $n$-th seed point $\mathbf{p}_n$ with selected rotation $\mathbf{R}_n$, each scene point $\mathbf{p}_i$ is transformed into the corresponding gripper frame:
\begin{equation}
\boldsymbol{\xi}_{n,i}
=
\mathbf{R}_n^\top(\mathbf{p}_i-\mathbf{p}_n).
\end{equation}
We define the gripper-aligned local support region as a cuboid in this gripper frame:
\begin{equation}
\mathcal{N}_n =
\left\{
i \ \middle|\ 
|\xi^x_{n,i}| \leq \frac{W_{\max}}{2},\
|\xi^y_{n,i}| \leq \frac{T_f}{2},\
0 \leq \xi^z_{n,i} \leq z_{\max}
\right\},
\end{equation}
where $W_{\max}$ is the maximum gripper width, $T_f$ is the fingertip thickness, and $z_{\max}$ is the maximum grouping extent along the depth direction.
Fig.~\ref{fig:grasp_representation}(b) illustrates this grouping region.

Unlike cylinder grouping~\cite{graspness,ma2023towards,anygrasp}, whose support region is mainly determined by the grasp center and approach direction, grasp-shaded grouping is conditioned on the full selected rotation, including the in-plane angle.
Different in-plane rotations therefore induce different cuboid support regions, enabling the model to aggregate candidate-specific image-informed features.

For each selected seed, we sample $K$ neighboring points from $\mathcal{N}_n$.
The grouped point features $\mathbf{F}^{\prime}_{\mathrm{group}} \in \mathbb{R}^{N \times K \times C}$ are aggregated by max pooling over the $K$ neighboring features, yielding $\mathbf{F}_{\mathrm{group}} \in \mathbb{R}^{N \times C}$.
From each grouped region, a three-layer operation head predicts grasp scores and corresponding widths for the $D$ depth anchors, yielding an output of size $N \times 2 \times D$.

\subsection{Optimization}
We optimize both backbones and prediction heads using four objectives: point-wise objectness loss ($L_o$), rotational graspness loss ($L_r$), grasp score loss ($L_s$), and grasp width loss ($L_\omega$).
The overall loss is
\begin{equation}
\label{eq:loss}
L = \lambda_o L_o + \lambda_r L_r + \lambda_s L_s + \lambda_\omega L_\omega,
\end{equation}
where $\lambda_o$, $\lambda_r$, $\lambda_s$, and $\lambda_\omega$ denote the corresponding loss weights.
The individual losses are:
\begin{align}
L_o &= \frac{1}{N_f} \sum_{i=1}^{N_f} L_{\mathrm{CE}}(o_i, \hat{o}_i), \\
L_r &= \frac{1}{NM} \sum_{n=1}^{N} \sum_{m=1}^{M} L_{\mathrm{reg}}(s_{n,m}, \hat{s}_{n,m}), \\
L_s &= \frac{1}{ND} \sum_{n=1}^{N} \sum_{d=1}^{D} L_{\mathrm{reg}}(q_{n,d}, \hat{q}_{n,d}), \\
L_\omega &= \frac{1}{ND} \sum_{n=1}^{N} \sum_{d=1}^{D} L_{\mathrm{reg}}(\omega_{n,d}, \hat{\omega}_{n,d}) \mathds{1}(\hat{q}_{n,d} > 0),
\end{align}
where $N_f$, $N$, $M$, and $D$ denote the numbers of full-scene points, selected seed points, joint view--in-plane rotation anchors, and depth anchors, respectively.
Here, $o_i$ denotes the predicted objectness score, $s_{n,m}$ the predicted rotational graspness score, and $q_{n,d}$ and $\omega_{n,d}$ the predicted grasp quality and gripper width for seed $n$ at depth anchor $d$.
The notation $\hat{\cdot}$ denotes the corresponding ground-truth value.
$L_{\mathrm{CE}}(\cdot)$ denotes cross-entropy loss, and $L_{\mathrm{reg}}(\cdot)$ denotes smooth-$L_1$ loss.
The width loss is applied only to grasp poses with positive ground-truth grasp scores.

\section{Experiments}
\label{sec:experiments}
We evaluate \methodname\ from five complementary perspectives: benchmark accuracy, the coupled effects of hierarchical image-to-3D feature injection and the rotational graspness model, cross-modal representation formation, robustness to point and RGB corruptions, and real-robot grasping.
We first report comparisons with state-of-the-art methods, then conduct Ablation studies and transfer analysis, and finally examine robustness, deployment cost, and real-world performance.

\subsection{Implementation Details}
\label{sec:implementation}
\noindent\textbf{Benchmark.} 
We evaluate our method using the GraspNet-1Billion dataset~\cite{graspnet}, which includes 100 training and 90 testing scenes, each captured from 256 views with two commercial-grade cameras, RealSense and Kinect.
The testing scenes are split into three categories: seen, similar, and novel, depending on whether the objects are included in the training data.

\noindent\textbf{Implementation.} 
Our method is implemented using PyTorch.
For scene-level grasp synthesis, we randomly sample $N_f=20{,}000$ points from the full scene point cloud as input and select $N=1{,}024$ seed points for grasp prediction.
Our image backbone is implemented as a ResNet34 encoder~\cite{he2016deep}, followed by a PSPNet~\cite{zhao2017pyramid} with four pooling layers of kernel sizes $(1\times1, 2\times2, 3\times3, 6\times6)$.
The image backbone produces a multi-scale image feature pyramid at strides $\{1,2,4,8,16\}$, and the channel dimension of the image features is $C_{\mathrm{img}}=64$.
MinkUNet14D~\cite{choy20194d} is used as our 3D backbone, producing image-informed 3D scene features with channel dimension $C=256$.
We implement hierarchical image-to-3D feature injection at encoder stages with strides $\{1,2,4,8,16\}$, matching the multi-scale image pyramid to the corresponding sparse 3D hierarchy.

The rotational graspness prediction stage employs an MLP with channel sizes $(256, 256, M)$, where $M=300\times12$ denotes the number of joint view--in-plane rotation anchors, consisting of 300 views and 12 in-plane rotations.
We set the quality threshold to $c=0.5$.
The grasp-shaded grouping stage samples $K=32$ neighboring points from each gripper-aligned cuboid region.
The cuboid dimensions are set to $W_{\max}=0.10$m, $T_f=0.02$m, and $z_{\max}=0.05$m, corresponding to the maximum gripper width, fingertip thickness, and maximum grouping extent along the depth direction, respectively.
For the operation head, we use an MLP with channel sizes $(256, 256, 2 \times D)$ to predict grasp scores and gripper widths for depth anchors, where $D=4$ is the number of depth anchors.
In loss design, we set $\lambda_o=1$, $\lambda_r=5$, $\lambda_s=5$, and $\lambda_\omega=1$.

\noindent\textbf{Training.} 
The network is trained on a single NVIDIA RTX3090 GPU using the AdamW optimizer~\cite{adamw} with a batch size of 6.
A cosine annealing learning rate is used, starting at $1.5 \times 10^{-3}$ and gradually decreasing to $10^{-5}$.
We train separate \methodname\ models for the two cameras for $20$ epochs.

\noindent\textbf{Evaluation.} 
We evaluate grasp accuracy using Average Precision (\textbf{AP}) following the GraspNet-1Billion protocol~\cite{graspnet,graspness}.
We compute $\textbf{AP}_{th}$ with the friction threshold $th$ varying from 0.2 to 1.2.
Non-Maximum Suppression (NMS) is applied before retrieving the \textbf{Top-50} highest-scoring grasps for evaluation, following~\cite{graspnet}.
For further details on the evaluation process, please refer to~\cite{graspnet}.

\begin{table}[!htbp]
	\centering
	\caption{Comparison with state-of-the-art methods on the GraspNet-1Billion dataset~\cite{graspnet}. Each entry reports RealSense/Kinect results.}
        \begin{threeparttable}
        \setlength{\tabcolsep}{3pt}
        \renewcommand{\arraystretch}{1.08}

\begin{tabularx}{\linewidth}{
@{}
>{\raggedright\arraybackslash}X
c
>{\centering\arraybackslash}p{1.85cm}
>{\centering\arraybackslash}p{1.85cm}
>{\centering\arraybackslash}p{1.85cm}
>{\centering\arraybackslash}p{1.85cm}
@{}
}
\Xhline{0.8pt}
\textbf{Model}
& \textbf{Input}
& \textbf{Seen}
& \textbf{Similar}
& \textbf{Novel}
& \textbf{Mean AP} \\
\hline

GPD~\cite{ten2017grasp}
& P
& 22.87/24.38
& 21.33/23.18
& 8.24/9.58
& 17.48/19.05 \\

PointNetGPD~\cite{liang2019pointnetgpd}
& P
& 25.96/27.59
& 22.68/24.38
& 9.23/10.66
& 19.29/20.88 \\

Fang~\textit{et al.}~\cite{graspnet}
& P
& 27.56/29.88
& 26.11/27.84
& 10.55/11.51
& 21.41/23.08 \\

RGB Matters~\cite{gou2021rgb}
& P+I
& 27.98/32.08
& 27.23/30.40
& 12.25/13.08
& 22.49/25.19 \\

Li~\textit{et al.}~\cite{li2021simultaneous}
& P
& 36.55/\nodata
& 28.36/\nodata
& 14.01/\nodata
& 26.31/\nodata \\

HGGD~\cite{chen2023efficient}
& P+I
& 59.36/60.26
& 51.20/48.59
& 22.17/18.43
& 44.24/42.43 \\

GSNet~\cite{graspness}
& P
& 67.12/63.50
& 54.81/49.18
& 24.31/19.78
& 48.75/44.15 \\

Ma~\textit{et al.}~\cite{ma2023towards}
& P
& 63.83/\nodata
& 58.46/\nodata
& 24.63/\nodata
& 48.97/\nodata \\

AnyGrasp~\cite{anygrasp}
& P
& 66.12/\nodata
& 56.09/\nodata
& 24.81/\nodata
& 49.01/\nodata \\

Wu~\textit{et al.}~\cite{EconomicGrasp}
& P
& 68.21/\nodata
& 61.19/\nodata
& 25.48/\nodata
& 51.63/\nodata \\

He~\textit{et al.}~\cite{he2024interest}
& P
& \nodata/66.77
& \nodata/54.99
& \nodata/23.52
& \nodata/48.43 \\

ZeroGrasp~\cite{zerograsp}
& P+I
& 70.53/\nodata
& 62.15/\nodata
& 26.46/\nodata
& 53.05/\nodata \\

Zhu~\textit{et al.}~\cite{zhu2026crossview}
& P+I
& 74.08/64.20
& 62.38/53.41
& 27.27/21.38
& 54.58/46.33 \\

FlexLoG~\cite{xie2026rethinking}
& P+I
& 72.81/\textbf{69.44}
& 65.21/\textbf{59.01}
& 30.04/23.67
& 56.02/50.67 \\

\hline
Ours
& P+I
& \textbf{77.52}/68.63
& \textbf{70.75}/58.05
& \textbf{32.74}/\textbf{25.49}
& \textbf{60.34}/\textbf{50.72} \\

\Xhline{0.8pt}
\end{tabularx}%
        \begin{tablenotes}
          \item \textit{P: point cloud; I: RGB image. Zhu~\textit{et al.}~\cite{zhu2026crossview} uses two RGB-D views.}
        \end{tablenotes}
        \end{threeparttable}
\label{tab:method_results}
\end{table}

\subsection{Comparison with State-of-the-Art Methods}

We strictly follow the evaluation protocol of GraspNet-1Billion~\cite{graspnet} to ensure fair comparisons.
Tab.~\ref{tab:method_results} summarizes the results on real-world scenes from the RealSense and Kinect cameras.
\methodname\ achieves the highest mean AP on both camera settings, reaching 60.34 on RealSense and 50.72 on Kinect.
On RealSense, it surpasses the strongest prior method, FlexLoG~\cite{xie2026rethinking}, by 4.32 mean AP and achieves the best performance on the Seen, Similar, and Novel splits.
On Kinect, \methodname\ is comparable to the strongest prior result in mean AP (50.72 vs.\ 50.67), while attaining the highest Novel-split AP of 25.49.
These results establish the overall competitiveness of \methodname\ across the two camera settings, with particularly strong performance on RealSense and the Novel split.
The contributions of hierarchical injection and rotational grasp modeling are isolated in the ablation studies below.
Qualitative comparisons are provided in the supplementary material.

\subsection{Ablation Studies and Representation Analysis}
\label{sec:ablations}
\rui{
Our analysis follows the two bottlenecks identified in Sec.~\ref{sec:introduction}: preserving multi-scale image context during hierarchical 3D representation formation and using the resulting image-informed features for fine-grained local grasp prediction.
We first evaluate the Rotational model and analyze what grasp information its pre-grouping representation retains.
We then compare alternative image-to-3D fusion strategies and examine how hierarchical injection shapes cross-modal representations.
Finally, we apply hierarchical injection to other grasp models while retaining their original prediction heads.
}

\begin{table}[!htbp]
	\centering
	\caption{Comparison of graspness models and image-to-3D fusion.}
    \begin{threeparttable}
    \begin{tabular}{cccccc}
\Xhline{0.6pt}
\multirow{2}*{\makecell[c]{\textbf{Graspness}\\\textbf{Model}}} &
\multirow{2}*{\makecell[c]{\textbf{Fusion}}} & \multicolumn{4}{c}{\textbf{AP}} \\
\cline{3-6}
& & \textbf{Seen} & \textbf{Similar} & \textbf{Novel} & \textbf{Mean}\\
\hline
View & None & 67.12 & 54.81 & 24.31 & 48.75 \\
View & Hierarchical & 71.97 & 58.37 & 25.50 & 51.95 \\
\hline
Rotational & None              & 71.61 & 62.96 & 30.30 & 54.95 \\
Rotational & Direct RGB$^{\ddag}$  &  71.86 & 61.84 & 28.85 & 54.18 \\
Rotational & Early             & 76.50 & 68.44 & 32.22 & 59.05 \\
Rotational & Late (Concat)     & 75.08 & 65.29 & 29.52 & 56.63 \\
Rotational & Late (Add)        & 76.91 & 67.21 & 31.62 & 58.58 \\
Rotational & Late (Gate)       & 74.73 & 64.84 & 30.40 & 56.66 \\
\hline
Rotational & Hierarchical  &  \textbf{77.52} & \textbf{70.75} & \textbf{32.74} & \textbf{60.34}\\
\Xhline{0.6pt}
\end{tabular}
    \begin{tablenotes}
      \item \textit{Note: Direct RGB $\ddag$ denotes concatenating RGB values \\ with 3D coordinates at the input.}
    \end{tablenotes}
    \end{threeparttable}
\label{tab:arch}
\end{table}

\noindent\textbf{Rotational Graspness Model.}
We first compare the conventional View model, which combines view graspness with cylinder grouping, against the Rotational model.
As shown in Tab.~\ref{tab:arch}, the Rotational model improves mean AP from 48.75 to 54.95 without image features and from 51.95 to 60.34 with hierarchical injection.
Notably, hierarchical injection improves the View model by 3.20 AP points but the Rotational model by 5.39 points, suggesting that the image-informed representation becomes more useful when grasp reasoning preserves the full gripper rotation.

We further decompose the Rotational model in Tab.~\ref{tab:grouping_nsample}.
Under hierarchical injection, replacing view graspness with rotational graspness while retaining conventional cylinder grouping increases mean AP from 51.95 to 56.87 (+4.92).
Conditioning the local support region on the selected full rotation through grasp-shaded grouping provides a further 3.47-point gain, reaching 60.34.
This decomposition indicates that the improvement is not solely due to changing the grouping region: resolving joint view--in-plane rotations already improves grasp prediction before the rotation-conditioned support is introduced.

We next vary the number of grouped samples $K$ to examine the sample efficiency of the two grouping strategies.
Reducing $K$ from 32 to 8 decreases mean AP of cylinder grouping by 4.74 points, whereas grasp-shaded grouping loses only 1.08 points.
At $K=16$, grasp-shaded grouping is already within 0.31 points of its $K=32$ result, and with only eight samples it exceeds cylinder grouping by 7.13 points.
These results show that grasp-shaded grouping provides more sample-efficient local aggregation under hierarchical injection.

\begin{table}[!htbp]
\centering
\caption{Sensitivity to the number of grouped local samples $K$ under hierarchical image-to-3D injection.
$\Delta$ denotes the mean AP change relative to $K=32$ for the same grouping type.}
\label{tab:grouping_nsample}
\begin{tabular}{lcccccc}
\toprule
\textbf{Grouping} & \textbf{$K$} & \textbf{Seen} & \textbf{Similar} & \textbf{Novel} & \textbf{Mean} & \textbf{$\Delta$ Mean} \\
\midrule
\multirow{4}{*}{Cylinder}
& 8  & 68.32 & 59.86 & 28.22 & 52.13 & -4.74 \\
& 16 & 71.46 & 62.91 & 30.19 & 54.85 & -2.02 \\
& 32 & 73.50 & 65.69 & 31.41 & 56.87 &  0.00 \\
& 64 & 74.02 & 66.04 & 31.67 & 57.24 & +0.38 \\
\midrule
\multirow{4}{*}{Grasp-shaded}
& 8  & 76.13 & 69.90 & 31.76 & 59.26 & -1.08 \\
& 16 & 77.21 & 70.36 & 32.51 & 60.03 & -0.31 \\
& 32 & 77.52 & 70.75 & 32.74 & 60.34 &  0.00 \\
& 64 & 77.57 & 70.83 & 32.83 & 60.41 & +0.07 \\
\bottomrule
\end{tabular}
\end{table}

\noindent\textbf{Pre-grouping Representation Analysis.}
To examine what graspness information is retained before local aggregation, we freeze the View and Rotational models and fit linear probes to their pre-grouping image-informed 3D features at identical query points.
We obtain \emph{view graspness} by averaging rotational graspness over in-plane rotations, and define the \emph{rotation residual} as each rotation's deviation from the view average.
These targets capture coarse view graspability and finer within-view rotation differences, respectively.
As shown in Tab.~\ref{tab:rotation_probe}, View Top-1 accuracy changes only from 3.80\% to 4.25\%, whereas rotation-residual Spearman correlation and Exact Top-1 accuracy improve from 0.311 to 0.367 and from 35.76\% to 39.15\%, respectively.
This contrast indicates that the main representational change occurs in within-view rotation information rather than generic view discrimination.
Rotation regret, which measures the graspness gap between the best and probe-selected rotations, also decreases from 0.2765 to 0.2516, showing that the Rotational representation better preserves the ranking among candidate in-plane rotations.
Together, these results are consistent with the intended role of rotational graspness: retaining grasp-quality distinctions among rotations before local grouping.
Further details are provided in the supplementary material.

\begin{table}[!htbp]
\centering
\caption{Representation analysis of pre-grouping image-informed 3D features.}
\label{tab:rotation_probe}
\begin{tabular}{llccc}
\toprule
Target
& Metric
& View
& Rotational
& $\Delta$ \\
\midrule
\makecell[l]{View\\graspness}
& Top-1 (\%) $\uparrow$
& 3.80
& \textbf{4.25}
& +0.45 \\
\midrule
\multirow{3}{*}{\makecell[l]{Rotation\\residual}}
& Spearman $\uparrow$
& 0.311
& \textbf{0.367}
& +0.056 \\
& Exact Top-1 (\%) $\uparrow$
& 35.76
& \textbf{39.15}
& +3.39 \\
& Regret $\downarrow$
& 0.2765
& \textbf{0.2516}
& $-0.0250$ \\
\bottomrule
\end{tabular}
\end{table}

\noindent\textbf{Comparison of Image-to-3D Fusion Scheme.}
Under the same rotational graspness model, we compare direct RGB input, early fusion, three late-fusion variants, and our hierarchical image-to-3D injection.
As reported in Tab.~\ref{tab:arch}, directly concatenating RGB values with point coordinates slightly reduces the mean AP from 54.95 to 54.18, indicating that isolated color measurements provide insufficient visual context.
Early fusion improves the mean AP to 59.05 by introducing contextualized image features before sparse 3D encoding, but exposes the 3D hierarchy to this evidence only once.
The late-fusion variants provide smaller gains because image and geometric representations are formed largely independently; among them, Late (Add) reaches 58.58 mean AP.
Hierarchical injection introduces aligned multi-scale image features at successive sparse-encoder stages and achieves the highest AP on all three test splits, improving the mean AP by 5.39 points.
These results indicate that image features are more effective when introduced before or throughout hierarchical 3D representation formation than when fused after the 3D representation has largely formed.
Further analyses in the supplementary material show that, among the learned RGB fusion variants, the advantage of hierarchical injection is not attributable to larger model capacity, or a greater optimization budget, while its computational cost remains comparable to the alternative fusion strategies.

\noindent\textbf{Cross-modal Correspondence under Hierarchical Injection.}
We extract sparse 3D features at voxel strides \texttt{s2/s4/s8} and the final backbone output, and pair them with RGB pyramid features \texttt{p1/p2/p4/p8} through pixel--point correspondence.
Centered Kernel Alignment (CKA)~\cite{kornblith2019similarity,raghu2021vision}
measures similarity in representational geometry and has been used for both layer-wise and cross-modal representation analysis.
Because CKA is symmetric, we complement it with bidirectional linear predictability~\cite{li2024vision,kapoor2025bridging}.
We fit linear mappings in both 3D$\rightarrow$image and image$\rightarrow$3D directions and report held-out $R^2$, measuring how much variance in one representation can be linearly predicted from the other.
Together, the two measures capture both geometric similarity and directional correspondence between the modalities.

\begin{figure}[t]
    \centering
    \includegraphics[width=\linewidth]{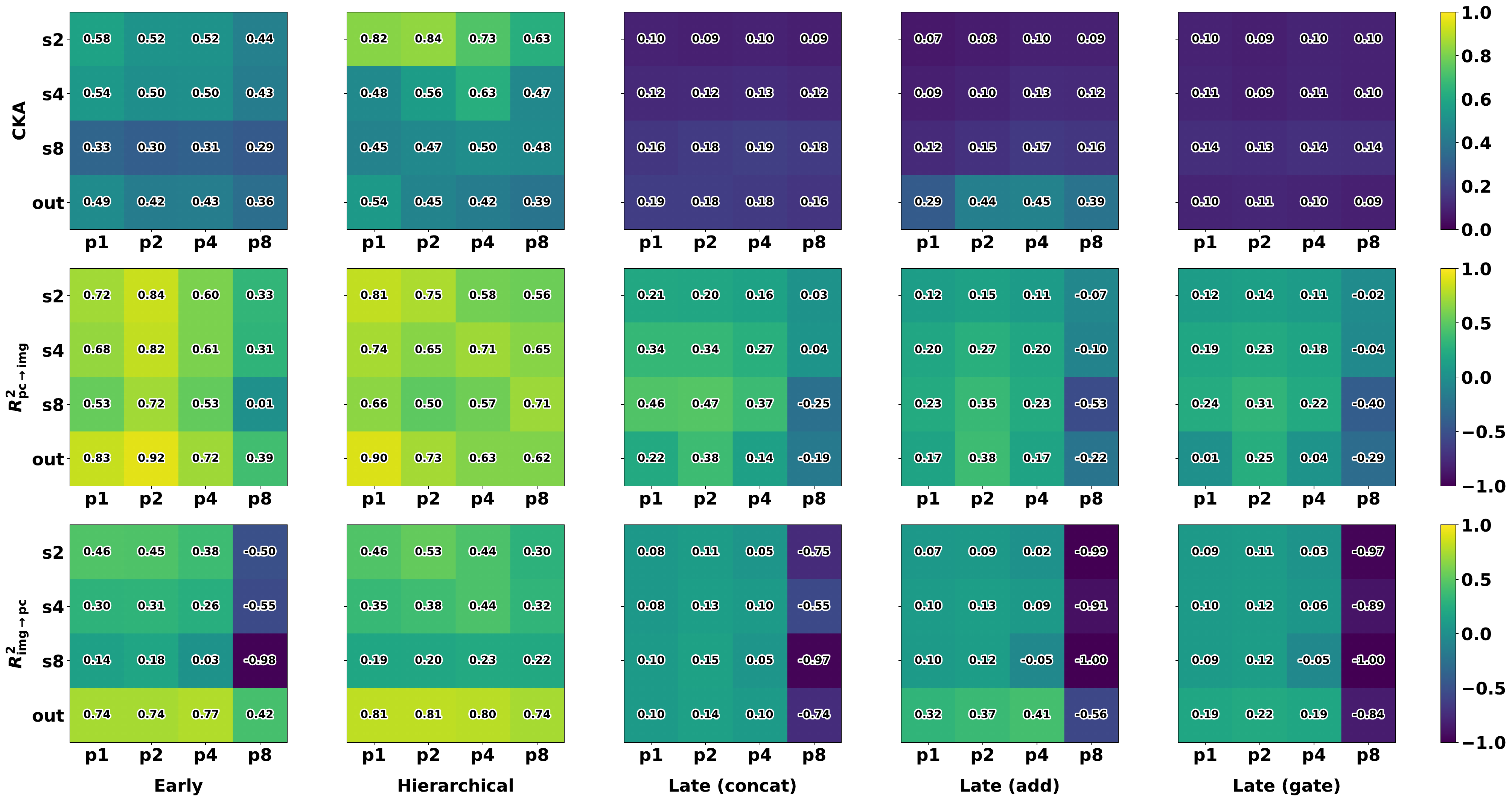}
    \caption{Cross-modal representation analysis on the test set. The top row reports CKA, while the middle and bottom rows report bidirectional linear-probe $R^2$ for 3D$\rightarrow$image and image$\rightarrow$3D prediction, respectively. Columns correspond to Early, Hierarchical, Late (Concat), Late (Add), and Late (Gate).}
     \label{fig:fusion_analy}
\end{figure}

As shown in Fig.~\ref{fig:fusion_analy}, hierarchical injection produces the strongest and most consistent cross-modal correspondence across network stages and image scales.
At the shallow \texttt{s2} stage, its CKA reaches 0.82/0.84 with \texttt{p1/p2}, compared with 0.58/0.52 for early fusion and approximately 0.10 for the late-fusion variants.
Hierarchical injection also maintains positive bidirectional $R^2$ over all stage--scale pairs.
Early fusion preserves strong fine-scale image information, but its image$\rightarrow$3D predictability decreases for coarse \texttt{p8} features as the 3D representation becomes deeper.
The late-fusion variants show little cross-modal correspondence at intermediate stages, consistent with the two modalities being processed independently before the final fusion operation.
Overall, these patterns are consistent with hierarchical injection preserving multi-scale visual evidence throughout 3D representation formation for subsequent grasp prediction.

\begin{table}[!htbp]
	\centering
	\caption{Transfer of hierarchical image-to-3D injection across different grasp methods.}
    \begin{tabular}{cccccc}
\Xhline{0.6pt}
\multirow{2}*{\makecell[c]{\textbf{Model}}} & 
\multirow{2}*{\makecell[c]{\textbf{Modality}}} & \multicolumn{4}{c}{\textbf{AP}} \\ 
\cline{3-6} 
& & \textbf{Seen} & \textbf{Similar} & \textbf{Novel} & \textbf{Mean}\\
\hline
GSNet~\cite{graspness} & P & 67.12 & 54.81 & 24.31 & 48.75 \\ 
+ Injection & P + I & 71.97 & 58.37 & 25.50 & 51.95 (\textbf{\textcolor{Orange}{+3.20}})\\ 
\hline
Ma~\textit{et al.}~\cite{ma2023towards} & P & 64.02 & 59.26 & 26.18 & 49.82 \\
+ Injection & P + I & 71.07 & 61.02 & 27.36 & 53.15 (\textbf{\textcolor{Orange}{+3.33}}) \\ 
\hline
\Xhline{0.6pt}
\end{tabular}
    \begin{tablenotes}
     \item \textit{Note: P: point cloud; I: RGB image.}
    \end{tablenotes}
\label{tab:generalization}
\end{table}

\noindent\textbf{Transfer across Grasp Backbones.}
We equip GSNet~\cite{graspness} and Ma~\textit{et al.}~\cite{ma2023towards} with the same image backbone and hierarchical image-to-3D injection while retaining their original grasp prediction heads.
All variants are retrained under the same settings.
As shown in Tab.~\ref{tab:generalization}, hierarchical injection improves the mean AP of GSNet from 48.75 to 51.95 and that of Ma~\textit{et al.} from 49.82 to 53.15, corresponding to gains of 3.20 and 3.33 points, respectively.
The consistent gains on these two representative backbones indicate that the benefit of hierarchical injection is not specific to \methodname's rotational graspness head.
The larger gain obtained by \methodname\ (5.39) is also consistent with the coupled ablation in Tab.~\ref{tab:arch}, where hierarchical injection provides a larger gain with the Rotational model than with the View model.

\subsection{Robustness under Point and RGB Corruptions}
\label{sec:robust_analysis}
Having established the benefit of hierarchical injection under standard observations, we examine controlled degradation of either modality.
For point corruptions, observed RGB is paired with a synthetic point cloud whose geometry alone is perturbed; for RGB corruptions, the naturally noisy observed point cloud is fixed while the corresponding observed RGB is perturbed.
\textbf{Point-only} denotes the same rotational graspness model without the image backbone or hierarchical image-to-3D injection.
We apply Gaussian noise, smoothing, and Perlin region dropout to the point cloud.
Perlin region dropout thresholds Perlin-noise masks~\cite{cheng20253d} to remove contiguous depth regions.
RGB corruptions include cutout, blur, brightness, and contrast.
All corruptions are applied only during evaluation; severity settings are provided in the supplementary material.

\begin{table}[!htbp]
\centering
\caption{Absolute mean AP under the strongest point corruptions.
AP retention (\%) relative to the uncorrupted synthetic-geometry reference is shown in parentheses.}
\label{tab:point_corruption}
\setlength{\tabcolsep}{4.0pt}
\begin{tabular}{lcccc}
\toprule
Method
& \makecell{Reference\\Mean AP}
& \makecell{Gaussian\\Noise}
& Smoothing
& \makecell{Perlin Region\\Dropout} \\
\midrule

GSNet~\cite{graspness}
& 55.92
& 9.25 (16.54)
& 49.19 (87.96)
& 50.13 (89.65) \\

Ma~\textit{et al.}~\cite{ma2023towards}
& 57.68
& 25.02 (43.37)
& 49.43 (85.71)
& 52.79 (\textbf{91.53}) \\

AnyGrasp~\cite{anygrasp}
& 60.42
& 30.45 (50.40)
& 50.27 (83.20)
& 53.04 (87.79) \\

Point-only
& 64.27
& 15.53 (24.16)
& 56.14 (87.35)
& 54.16 (84.26) \\

Ours
& \textbf{67.04}
& \textbf{39.28} (\textbf{58.60})
& \textbf{59.11} (\textbf{88.17})
& \textbf{61.26} (91.38) \\

\bottomrule
\end{tabular}
\end{table}

\noindent\textbf{Robustness under Point Corruptions.}
Tab.~\ref{tab:point_corruption} reports both absolute AP and AP retention under the strongest corruption of each type.
The reference AP is measured using observed RGB paired with the uncorrupted synthetic point cloud, while retention normalizes the corrupted AP by this reference.
\methodname\ achieves the highest absolute AP under all three corruption types and the highest retention under Gaussian noise and smoothing; Ma~\textit{et al.} is marginally higher in retention under Perlin region dropout (91.53\% vs.\ 91.38\%).

The paired comparison with Point-only isolates the contribution of hierarchical image-to-3D injection.
Under the strongest Gaussian noise, hierarchical injection raises absolute AP from 15.53 to 39.28, a 23.75-point gain, and increases retention from 24.16\% to 58.60\%.
Under Perlin region dropout, it raises AP from 54.16 to 61.26 and retention from 84.26\% to 91.38\%, indicating that dense image evidence compensates for missing point regions.
The smaller improvement under smoothing, from 56.14 to 59.11 AP, suggests that image evidence provides less compensation when local metric contact geometry is directly removed by over-smoothing.
Together, these results indicate that hierarchical injection is particularly effective under coordinate noise and structured point loss.
Full AP curves across corruption levels are provided in the supplementary material.

\begin{table}[!htbp]
\centering
\caption{Utility under clean and RGB-corrupted observations. Parentheses show AP differences over Point-only under the same observation condition.}
\label{tab:rgb_utility_sensitivity}
\begin{tabular}{lcc}
\toprule
Variant
& Clean AP $\uparrow$
& RGB-corrupted AP $\uparrow$ \\
\midrule
Point-only
& 54.95 $(+0.00)$
& 54.95 $(+0.00)$ \\

Early
& 58.96 $(+4.01)$
& 56.47 $(+1.51)$ \\

Late (Concat)
& 56.52 $(+1.56)$
& 54.34 $(-0.61)$ \\

Late (Add)
& 58.52 $(+3.56)$
& 56.55 $(+1.60)$ \\

Late (Gate)
& 56.59 $(+1.64)$
& 53.06 $(-1.89)$ \\

Hierarchical (Ours)
& \textbf{60.33} $(\mathbf{+5.38})$
& \textbf{57.28} $(\mathbf{+2.32})$ \\
\bottomrule
\end{tabular}
\end{table}

\noindent\textbf{Utility under RGB Corruptions.}
Tab.~\ref{tab:rgb_utility_sensitivity} reports clean image utility and absolute performance under RGB corruption, averaged over all corruption types and severity levels.
Hierarchical injection achieves the highest clean AP, improving over Point-only by 5.38 AP points, and remains best after corruption with a 2.32-point advantage.
Late (Add) is less affected by RGB corruption, but its lower clean and corrupted AP indicate more conservative use of image evidence.
Late (Concat) and Late (Gate) fall below Point-only after corruption, showing that unreliable image features can become negative evidence.

The corruption-wise results show that cutout causes by far the largest degradation, whereas blur, brightness, and contrast have substantially smaller effects.
Detailed results, visualizations, and analysis are provided in the supplementary material.

\noindent\textbf{Inference Efficiency and Cost.}
We report the grasp-model inference latency and resource cost in Tab.~\ref{tab:deployment_cost}.
All measurements are conducted on a server equipped with an Intel(R) Xeon(R) Gold 6226R CPU @ 2.90\,GHz and a single NVIDIA GeForce RTX 3090.
All models process 20,000 scene points as input, while all other method-specific hyperparameters and inference settings follow the configurations reported in their respective papers.

GSNet and AnyGrasp require 33.82\,ms and 44.55\,ms per sample, respectively.
Despite additionally extracting and hierarchically injecting RGB features, \methodname\ requires 71.74\,ms, remaining within a practical inference range and substantially faster than Ma~\textit{et al.} (784.05\,ms).

\methodname\ contains 44.14M parameters and requires 304.47 profiled GFLOPs per sample.
Its peak inference and training memory are 539.86\,MiB and 16.18\,GiB, respectively, while training for 20 epochs requires 1.78 GPU-days.
Detailed controlled comparisons of parameters, computation, latency, memory, and training cost across different fusion variants are provided in the supplementary material.

\begin{table}[!htbp]
\centering
\caption{Inference latency comparison and deployment cost of \methodname.}
\begin{tabular}{@{}c@{\hspace{14pt}}c@{}}

\begin{tabular}[t]{l c}
\multicolumn{2}{c}{(a) Model latency comparison} \\[2pt]
\toprule
Model & Time (ms) \\
\midrule
GSNet~\cite{graspness}              & 33.82  \\
AnyGrasp~\cite{anygrasp}            & 44.55  \\
Ma~\textit{et al.}~\cite{ma2023towards} & 784.05 \\
Ours                        & 71.74  \\
\bottomrule
\end{tabular}
&
\begin{tabular}[t]{l c}
\multicolumn{2}{c}{(b) Resource cost} \\[2pt]
\toprule
Item & Value \\
\midrule
Parameters          & 44.14M     \\
Profiled GFLOPs     & 304.47     \\
Infer. peak VRAM    & 539.86 MiB \\
Train peak VRAM     & 16.18 GiB  \\
Training cost       & 1.78 GPU-days \\
\bottomrule
\end{tabular}
\end{tabular}
\label{tab:deployment_cost}
\end{table}

\begin{figure}[t]
    \centering
    \includegraphics[width=\linewidth]{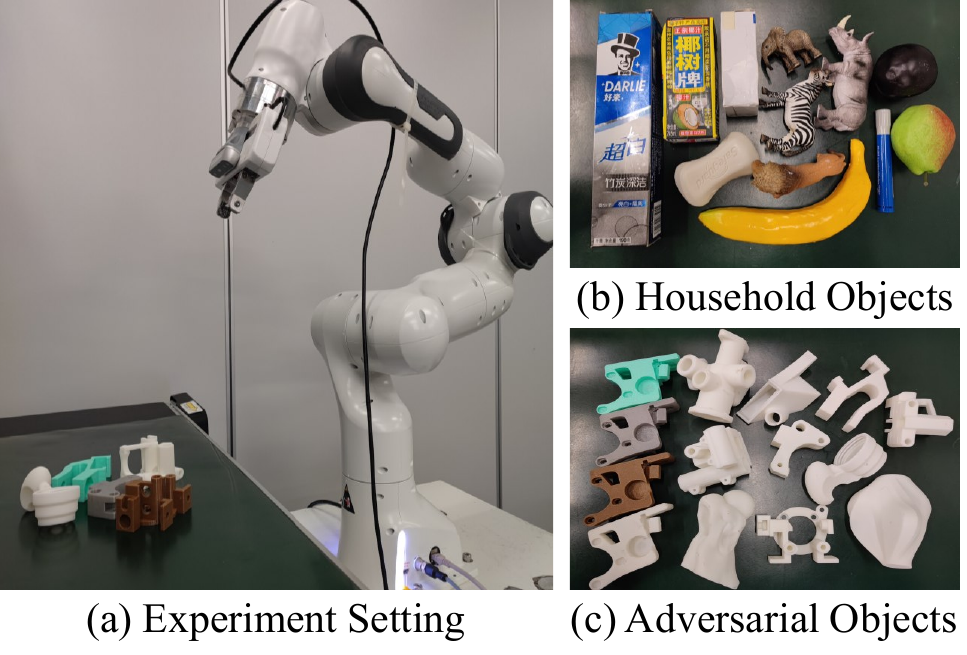}
    \caption{Our setup of the real-world evaluation.}
\label{fig:real_setting}
\end{figure}

\begin{table}[!htbp]
	\centering
	\caption{Results of real-world evaluation}
    \begin{tabular}{lcccc}
\Xhline{0.6pt}
\hline
\multirow{2}{*}{\textbf{Model}}
& \multicolumn{2}{c}{\textbf{Household}}
& \multicolumn{2}{c}{\textbf{Adversarial}} \\
\cline{2-5}
& \textbf{Succ./TA} & \textbf{SR}(\%)
& \textbf{Succ./TA} & \textbf{SR}(\%) \\
\hline
Point-only
& 108/140 & 77.14
& 102/143 & 71.33 \\
\hline
AnyGrasp~\cite{anygrasp}
& 110/133 & 82.71
& 108/148 & 72.97 \\
\hline
Ours
& \textbf{116/138} & \textbf{84.06}
& \textbf{114/140} & \textbf{81.43} \\
\hline
\Xhline{0.6pt}
\end{tabular}
\label{tab:real_evaluation}
\end{table}

\subsection{Real-world Evaluation}
\label{sec:real_world}
To assess the practical effectiveness of our method, we conducted real-world experiments using a robotic system based on a 7-DOF Franka Panda robot arm equipped with a Franka Hand.
An Intel RealSense D415 camera is mounted directly on the gripper for visual perception. 
The test set includes 22 object types from the GraspNet-1Billion dataset~\cite{graspnet}, split into two groups: \textbf{Household} (seen and similar objects) and \textbf{Adversarial} (novel objects with multiple color variants)~\cite{wang2019adversarial}. 
Each group is evaluated over 20 trials, each conducted in a cluttered scene containing 5--6 randomly selected objects.
A detailed illustration of our setup is shown in Fig.~\ref{fig:real_setting}.

We evaluate the Point-only baseline, \methodname, and AnyGrasp~\cite{anygrasp} using the same post-processing steps. We record Total Attempts (TA) and Success Rate (SR), where SR is the proportion of successful grasps among all attempts. Further details on the robot setup, exception handling, and post-processing are provided in the supplementary material.

As shown in Tab.~\ref{tab:real_evaluation}, \methodname\ achieves success rates of 84.06\% on Household objects and 81.43\% on Adversarial objects.
It improves over the Point-only baseline by 6.92 and 10.10 percentage points, respectively, providing real-robot evidence for the benefit of hierarchical image-to-3D injection under the same rotational graspness model.
Compared with AnyGrasp~\cite{anygrasp}, \methodname\ improves by 1.35 percentage points on Household objects and 8.46 percentage points on Adversarial objects.
The larger gain on Adversarial objects is consistent with the corruption analysis in Sec.~\ref{sec:robust_analysis}, where image evidence provides complementary information when point observations are degraded.

\begin{figure}[t]
    \centering
    \includegraphics[width=\linewidth]{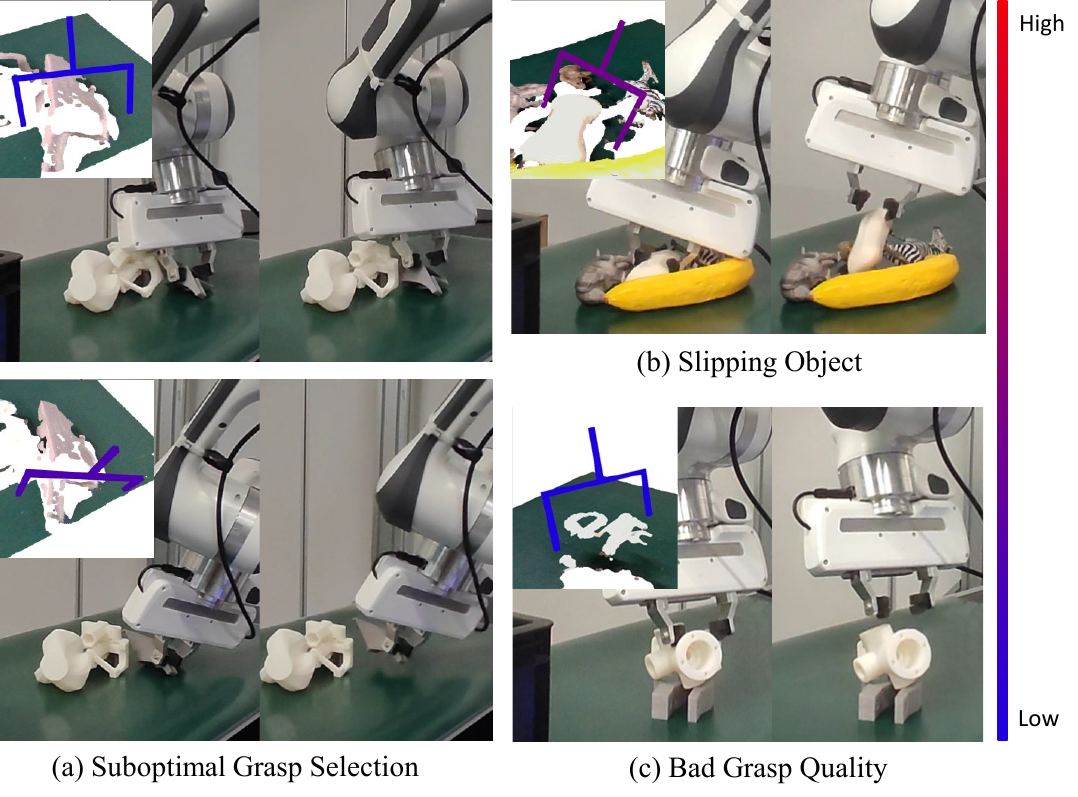}
    \caption{Failure modes in our real-world experiments.}
     \label{fig:failre_mode}
\end{figure}

\subsection{Failure Mode Analysis}
In our real-world evaluations, we observe three representative failure cases: (i) suboptimal grasp selection after the execution-angle constraint, (ii) object slippage, and (iii) inaccurate grasp depth. 
The first occurs when a high-quality prediction is filtered by the angle constraint, leaving an easier-to-execute but less reliable candidate.
Fig.~\ref{fig:failre_mode}(a) illustrates such a case. 
The bottom row of Fig.~\ref{fig:failre_mode}(a) shows another challenging case in which the model assigns high confidence to a grasp on a small object region.
The second failure mode occurs on low-friction objects, such as soap (Fig.~\ref{fig:failre_mode}(b)), which may slip from the gripper during execution. 
The third example is associated with an inaccurate predicted grasp depth, which can prevent the gripper from achieving force closure (Fig.~\ref{fig:failre_mode}(c)). 
Additional execution examples are provided in the supplementary material.

\section{Conclusion}
\label{sec:conclusion}
This work investigates how image information can be more effectively exploited for scene-level 6-DoF grasp synthesis.
\methodname\ combines hierarchical image-to-3D feature injection with a rotational graspness model, enabling multi-scale image context to participate throughout 3D representation formation and allowing the selected grasp rotation to guide local feature aggregation.
Ablation and representation analyses show that these designs improve image--3D correspondence and preserve more grasp-quality information among candidate rotations.
\methodname\ achieves 60.34 mean AP on the RealSense setting of GraspNet-1Billion and improves real-robot grasp success over AnyGrasp by 8.46 percentage points on challenging objects.
The consistent gains obtained when applying hierarchical injection to other grasp models further suggest that this image-to-3D interface can benefit broader scene-level grasp pipelines.

Several limitations remain.
The current model relies on accurate image--point correspondence and lacks an explicit fallback when RGB information becomes unreliable.
The additional image backbone also introduces extra computation.
Future work will explore reliability-aware and more efficient image-to-3D fusion, execution-aware grasp scoring, and online feedback for avoiding repeated failures.

\section*{CRediT authorship contribution statement}
\textbf{Rui Cao:}
Conceptualization, Methodology, Software, Validation, Formal analysis,
Investigation, Visualization, Writing - Original Draft,
Writing - Review \& Editing;
\textbf{Jia-Hui Pan:}
Methodology, Validation, Investigation, Writing - Review \& Editing;
\textbf{Yichuan Li:}
Software, Validation;
\textbf{Yunfan Yu:}
Validation;
\textbf{Chi-Wing Fu:}
Supervision, Writing - Review \& Editing;
\textbf{Jiangliu Wang:}
Writing - Review \& Editing;
\textbf{Yun-Hui Liu:}
Supervision, Project administration, Resources, Funding acquisition.

\section*{Acknowledgments}
This work is supported in part by Research Grants Council (RGC) of Hong Kong under Grant 14207324, in part by the HK RGC AoE under AoE/E-407/24-N, in part by the InnoHK initiative of the Innovation and Technology Commission of the Hong Kong Special Administrative Region Government via the Hong Kong Centre for Logistics Robotics, and in part by the CUHK T Stone Robotics Institute.

\section*{Declaration of competing interest}
The authors declare that they have no known competing ﬁnancial interests or personal relationships that could have appeared to inﬂuence the work reported in this paper.

\section*{Data availability}
The code, trained models, and data associated with this study will be made publicly available upon publication.

\section*{Declaration of generative AI and AI-assisted technologies in the manuscript preparation process}
During the preparation of this work, the authors used ChatGPT (OpenAI) to assist with language editing, proofreading, and \LaTeX{} formatting. After using this tool, the authors reviewed and edited the content as needed and take full responsibility for the content of the published article.

 
%

\bibliographystyle{elsarticle-num}
\bibliography{reference}

\end{document}